\documentclass{article}

\usepackage{microtype}
\usepackage{graphicx}
\usepackage{subfigure}
\usepackage{booktabs} 

\usepackage{hyperref}
\usepackage{bm} 
\usepackage{tabularx}

\newcolumntype{Y}{>{\centering\arraybackslash}X}
\newcolumntype{Z}{>{\centering\let\newline\\\arraybackslash\hspace{0pt}}X}
\newcolumntype{L}{>{\RaggedRight\hangafter=1\hangindent=0em}X}

\usepackage{color}
\usepackage{multirow}
\usepackage[misc]{ifsym}
\usepackage{balance}

\usepackage[accepted]{icml2024}

\usepackage{amsmath}
\usepackage{amssymb}
\usepackage{mathtools}
\usepackage{amsthm}

\usepackage[capitalize,noabbrev]{cleveref}

\theoremstyle{plain}

\theoremstyle{definition}

\theoremstyle{remark}

\usepackage[textsize=tiny]{todonotes}

\icmltitlerunning{Revisiting Context Aggregation for Image Matting}

\begin{document}

\twocolumn[
\icmltitle{Revisiting Context Aggregation for Image Matting  }




\begin{icmlauthorlist}
\icmlauthor{Qinglin Liu}{yyy}
\icmlauthor{ Xiaoqian Lv}{yyy}
\icmlauthor{Quanling Meng}{yyy}
\icmlauthor{Zonglin Li}{yyy}
\icmlauthor{Xiangyuan Lan}{pcs}
\icmlauthor{Shuo Yang}{yyy}
\icmlauthor{Shengping Zhang}{yyy}
\icmlauthor{Liqiang Nie}{yyy}

\end{icmlauthorlist}

\icmlaffiliation{yyy}{Harbin Institute of Technology}
\icmlaffiliation{pcs}{Peng Cheng Laboratory}

\icmlcorrespondingauthor{Shengping Zhang}{s.zhang@hit.edu.cn}

\icmlkeywords{Machine Learning, ICML}

\vskip 0.3in
]



\printAffiliationsAndNotice{}  

\begin{abstract}
Traditional studies emphasize the significance of context information in improving matting performance.
Consequently, deep learning-based matting methods delve into designing pooling or affinity-based context aggregation modules to achieve superior results.
However, these modules cannot well handle the context scale shift caused by the difference in image size during training and inference, resulting in matting performance degradation.
In this paper, we revisit the context aggregation mechanisms of matting networks and find that a basic encoder-decoder network without any context aggregation modules can actually learn more universal context aggregation, thereby achieving higher matting performance compared to existing methods.
Building on this insight, we present AEMatter, a matting network that is straightforward yet very effective.
AEMatter adopts a Hybrid-Transformer backbone with appearance-enhanced axis-wise learning (AEAL) blocks  to build a basic network with strong context aggregation learning capability.
Furthermore, AEMatter leverages a large image training strategy to assist the network in learning context aggregation from data.
Extensive experiments on five popular matting datasets demonstrate that the proposed AEMatter outperforms state-of-the-art matting methods by a large margin.
\end{abstract}

\begin{figure}[!t]
    \begin{center}
\subfigure[Basic Encoder-Decoder Matting Network]{
\label{fig:basicm}
\includegraphics[width=0.98\linewidth]{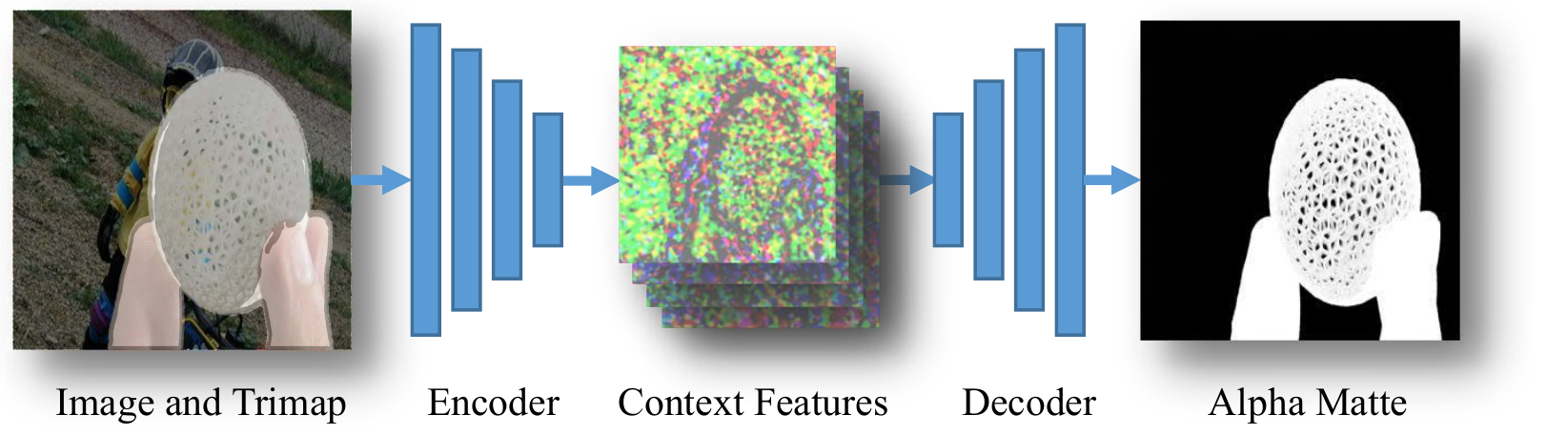}}
\vspace{-7pt}
\subfigure[Pooling-based Context Aggregation Module]{
\label{fig:pla}
\includegraphics[width=0.98\linewidth]{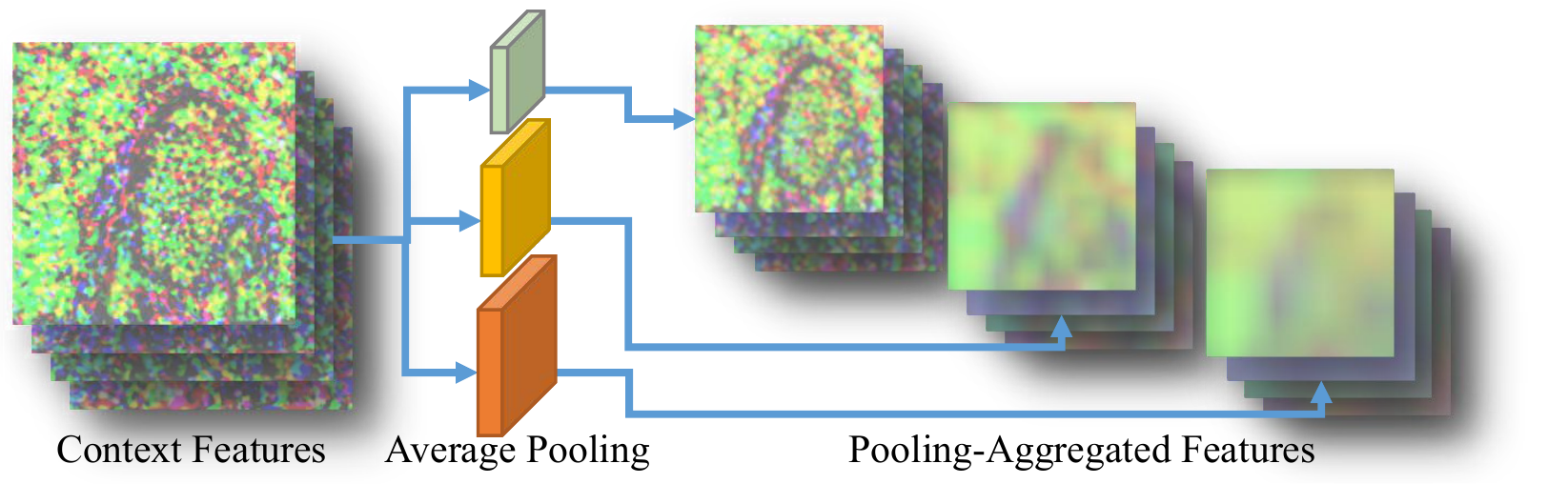}}
\vspace{-7pt}
\subfigure[Affinity-based Context Aggregation Module]{
\label{fig:afa}
\includegraphics[width=0.98\linewidth]{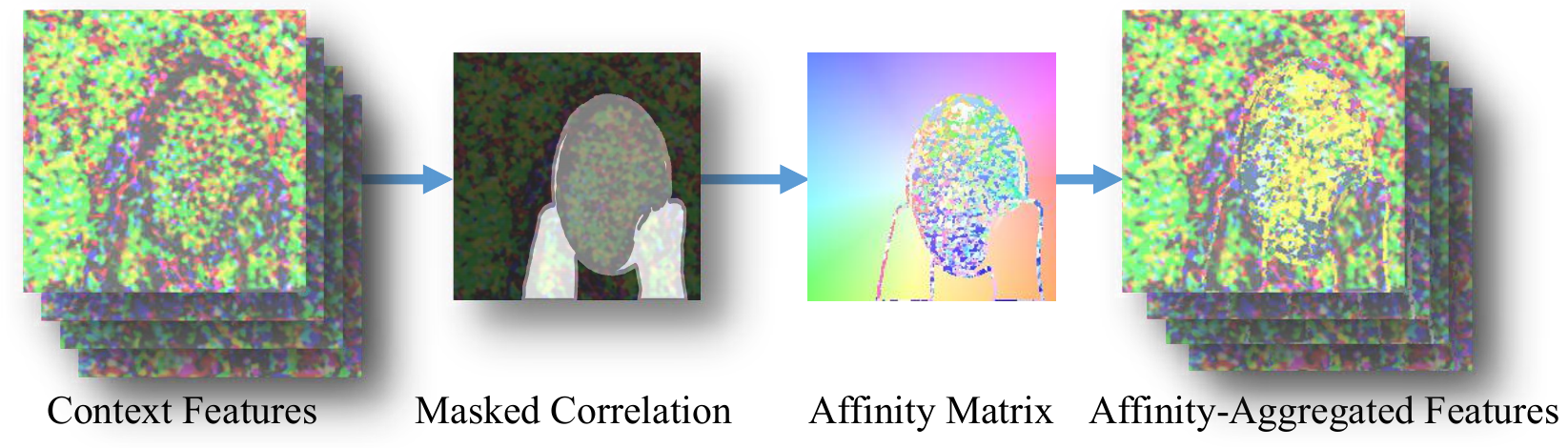}}
\vspace{-7pt}
    \end{center}
    \caption{Illustration of a basic matting network and context aggregation modules. (a) The basic matting network uses an encoder to extract context features from inputs, and a decoder to predict alpha mattes. Our AEMatter also follows this scheme. (b) Pooling-based context aggregation module uses pooling operations to aggregate contexts from  surrounding regions. (c) Affinity-based context aggregation module uses affinity operations to aggregate contexts from globally related regions.}
    \vspace{-0.28cm}
    \label{fig:overs}
\end{figure}

\section{Introduction}
\label{sec:intro}
Natural image matting is a classic problem that involves estimating the alpha matte of the foreground in a given image.
This technology has numerous real-world applications, such as image editing~\cite{2009Sketch2Photo,2017Robust} and film post-production~\cite{2015Integrated,Wang_2021_ICCV}. 
Formally, a given image $\bm{I}$ can be represented as a combination of a foreground $\bm{F}$ and background $\bm{B}$ as
\begin{equation}
\label{sca}
{{I}}_i = {\alpha}_i {F}_i + (1-{ \alpha}_i ){B}_i 
\end{equation}
where ${\alpha}_i$ is the alpha matte at pixel $i$.
Therefore, matting involves the challenge of regressing alpha matte $\bm{\alpha}$ based on image $\bm{I}$. 
This process not only necessitates distinguishing between foreground and background but also determining the weights of the foreground, making  it an intricate task.

To address the matting challenge, early researchers~\cite{berman1998method,ruzon2000alpha,grady2005random,levin2008a} explore to estimate alpha mattes based on location and color similarity or by propagating color information within a local region. 
To improve matting performance,  context aggregation technologies, such as global sampling or non-local propagation, are developed to leverage context information away from the foreground boundaries.
Recently, deep learning-based methods~\cite{xu2017deep,lu2019indices} employ basic encoder-decoder networks to extract context features from input data and estimate alpha mattes, as depicted in Figure~\ref{fig:overs}(a). Due to the formidable learning capability of neural networks, these methods outperform traditional matting methods by a substantial margin.
To further improve prediction accuracy, researchers emulate traditional methods in designing context aggregation modules to effectively exploit context information~\cite{li2020natural,forte2020fbamatting}. These modules adopt pooling or affinity based operations, as illustrated in Figures~\ref{fig:overs}(b) and \ref{fig:overs}(c), to aggregate context information.
However, it is rarely acknowledged that these modules cannot well handle the context scale shift caused by the difference in image size during training and inference, resulting in matting performance degradation.

In this paper, we revisit the context aggregation mechanisms of matting networks to inspire future research on high-performance matting methods.
Specifically, we first evaluate existing matting networks, revealing that networks with context aggregation modules usually exhibit more errors when inferring on larger images, compared to networks without such modules. 
This observation underscores that while context aggregation modules can effectively aggregate contexts, their sensitivity to context scale restricts their universality.
Subsequently, our assessment extends to basic encoder-decoder networks, where we observe their impressive performance. 
These results suggest that basic networks possess the capability to aggregate contexts for high-performance matting.
Further exploration reveals that enhancing context aggregation capability can be achieved through training with large image patches and incorporating network layers with a larger receptive field.
Building on these insights, we introduce AEMatter, a matting network that is both simpler and more powerful than existing methods. 
AEMatter adopts a Hybrid-Transformer backbone and integrates appearance-enhanced axis-wise learning (AEAL) blocks  to build a basic network with strong context aggregation learning capability.
Furthermore, AEMatter employs a large image training strategy to facilitate the network in learning context aggregation.
Extensive experiments on five matting datasets demonstrate that  AEMatter outperforms state-of-the-art methods by a large margin.

To summarize, the contributions of this paper are as follows:

\begin{itemize}

    \item We pioneer an experimental analysis to evaluate the effectiveness and mechanisms of context aggregation modules within existing matting networks. 
    Our findings reveal that while context aggregation modules can effectively aggregate contexts, their sensitivity to the context scale restricts their universality.
    
    \item We empirically find that basic encoder-decoder matting networks can learn to aggregate contexts for high-performance matting. Moreover, we demonstrate that this capability can be enhanced through training with large image patches and the adoption of network layers with a larger receptive field.

    \item We introduce AEMatter, a straightforward yet effective matting network that expands the receptive field with appearance-enhanced axis-wise learning (AEAL) blocks and is trained using large image patches.  Experimental results demonstrate that  AEMatter significantly outperforms state-of-the-art methods. 
\end{itemize}

\section{Related Work}
\noindent \textbf{Traditional matting methods.}
Traditional matting methods can be categorized into two approaches: sampling-based methods and propagation-based methods.
Sampling-based methods involve sampling candidate foreground and background colors for pixels in unknown regions to estimate the alpha matte. Bayesian Matting\cite{chuang2001a} models foreground and background colors with a Gaussian distribution and incorporates spatial location information to enhance accuracy. Global Matting~\cite{he2011a} takes a different approach by sampling pixels in all known regions to prevent information loss and improve robustness.
Propagation-based methods rely on the assumption that foreground and background colors exhibit smoothness in local regions for alpha matte estimation. Poisson Matting~\cite{sun2004poisson} utilizes boundary information from trimap to solve the Poisson equation, making it capable of estimating the alpha matte even with a rough trimap. Closed-form matting~\cite{levin2008a} introduces a color-line assumption and provides a closed-form solution for estimation. 

\noindent \textbf{Deep learning-based matting methods.}
Deep learning-based methods train the networks on image matting datasets to estimate the alpha matte.
Early methods~\cite{xu2017deep,lu2019indices} typically employ a basic encoder-decoder network for matting. 
DIM~\cite{xu2017deep} introduced a refinement module to the decoder to improve the performance.
IndexNet~\cite{lu2019indices} retains the indices of the downsampled features for improving the gradient accuracy.
Recent advancements in deep image matting methods have designed pooling-based or affinity-based context aggregation modules to refine context features and adopt other techniques to improve performance. 
Pooling-based methods~\cite{forte2020fbamatting,yu2020mask,sun2021sim,liu2021lfpnet,park2022matteformer,cai2022TransMatting} use average pooling to aggregate contexts from surrounding regions for context feature refinement.
FBAMatting~\cite{forte2020fbamatting} adopts pyramid pooling module (PPM)~\cite{Zhao2016Pyramid} and introduces the groupnorm~\cite{wu2018group} and weight standardization~\cite{weightstandardization} tricks to improve the matting performance.
MGMatting~\cite{yu2020mask} adopts ASPP and designs a progressive refinement decoder to estimate fine alpha mattes from coarse segmentation.
MatteFormer~\cite{park2022matteformer} proposes a trimap-guided token pooling module and adopts the Swin-Tiny~\cite{liu2021Swin} backbone to improve the prediction.
Affinity-based methods~\cite{li2020natural,yu2020high,Yu_2021_ICCV,dai2022boosting} use the masked correlation to construct an affinity matrix and enhance the context features with the contexts from globally related regions. 
GCAMatting~\cite{li2020natural} adopts the guided context attention module to improve the prediction in the transparent region.
TIMI-Net~\cite{Liu_2021_ICCV} proposes a tripartite information module and multi-branch architecture to improve predictions.
\begin{figure}[!t]
    \begin{center}
    \includegraphics[width=0.48\linewidth]{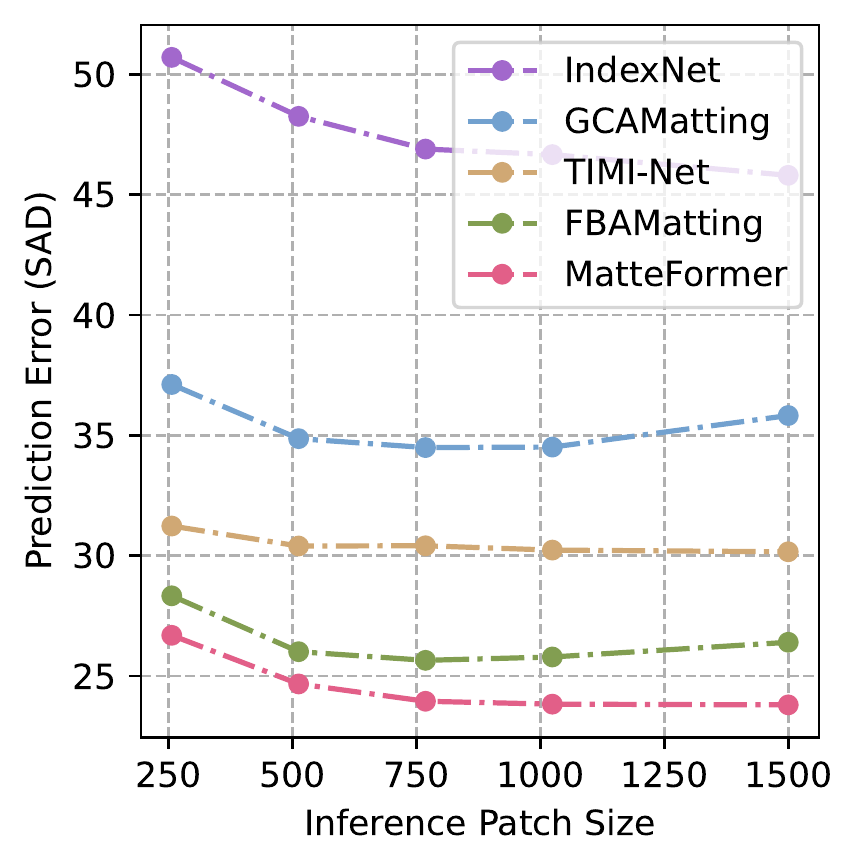}
    \includegraphics[width=0.48\linewidth]{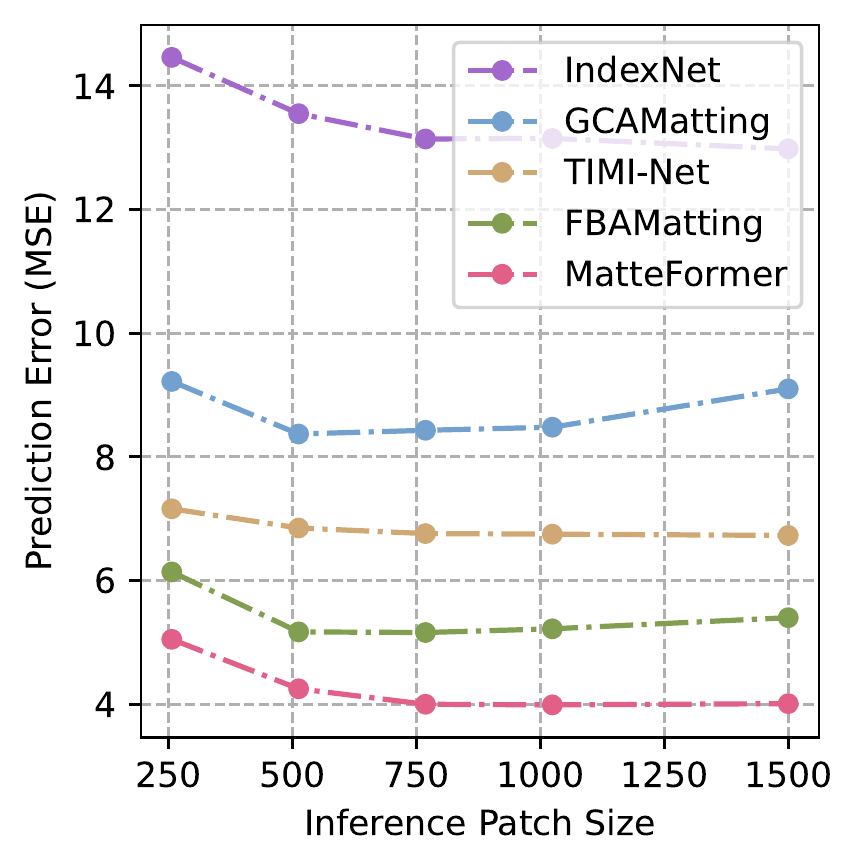}
    \end{center}
    \caption{  \textbf{Inference Patch Size} vs \textbf{Prediction Errors}. As the inference patch size increases, the prediction errors of the compared matting methods first decrease and then show different trends. }
    \label{fig:pvse}
\end{figure}

\section{Empirical Study}
In this section, we perform experimental analyses on existing matting networks and basic encoder-decoder matting networks to explore the context aggregation mechanisms of matting networks and identify the key factors contributing to the performance of matting networks.

\subsection{Exploring Existing Matting Networks}
We assess the performance and robustness of existing matting networks, observing that both the encoder-decoder and the context aggregation module within these networks can effectively aggregate contexts for matting. Nevertheless, the sensitivity of context aggregation modules to context scale restricts their universality.

\begin{table}[!t]
  \centering
  \caption{Comparison of state-of-the-art matting methods trained on Adobe Composition-1K using image patches of different sizes. }
\resizebox{1\linewidth}{!}{
    \begin{tabular}{l|c|c|c|c|c}
    \toprule
    Method & Patch Size & SAD   & MSE   & Grad  & Conn \\
    \midrule
    IndexNet~\cite{lu2019indices} & 256 & 38.52  & 8.74  & 18.02  & 36.43  \\
    IndexNet~\cite{lu2019indices} & 512 & 33.64  & 7.05  & 14.35  & 30.21 \\
    IndexNet~\cite{lu2019indices} & 768 & 31.12  & 6.40  & 12.83  & 27.63  \\
    IndexNet~\cite{lu2019indices} & 1024 & 30.91 & 6.73  & 13.72  & 27.17  \\
    \midrule
    FBAMatting~\cite{forte2020fbamatting} & 256 & 43.18  & 10.41   & 21.13  &42.39 \\
    FBAMatting~\cite{forte2020fbamatting} & 512 & 33.36  & 7.26   & 15.75  & 29.84 \\
    FBAMatting~\cite{forte2020fbamatting} & 768 & 29.89  & 5.73   & 14.05  & 26.18 \\
    FBAMatting~\cite{forte2020fbamatting} & 1024 & 30.76  & 5.74   & 15.19  & 27.03 \\
    \midrule
    MatteFormer~\cite{park2022matteformer} & 256 & 28.52  & 5.51  & 12.00  & 24.06 \\
    MatteFormer~\cite{park2022matteformer} & 512 & 23.61  & 3.78  & 9.23  &18.52 \\
    MatteFormer~\cite{park2022matteformer} & 768 &22.78  & 3.59  & 8.38  & 17.50 \\
    MatteFormer~\cite{park2022matteformer} & 1024 &23.68 & 3.62  &8.81  & 18.66 \\
    \bottomrule
    \end{tabular}
    }
  \label{tab:res}%
\end{table}%

\noindent \textbf{Patch-based Inference.}
Existing matting networks usually include an encoder-decoder network with a context aggregation module. The context aggregation modules, built with hard-crafted structures, are considered to exhibit better context aggregation capability across images of various sizes compared to the encoder-decoder network. 
To validate this understanding, we conduct a patch-based inference evaluation for existing matting networks.
We evaluate existing matting methods, including IndexNet~\cite{lu2019indices} without a context aggregation module and GCAMatting~\cite{li2020natural}, TIMI-Net~\cite{Liu_2021_ICCV}, FBAMatting~\cite{forte2020fbamatting}, and MatteFormer~\cite{park2022matteformer} with a context aggregation module.
The evaluation was conducted on image patches of varying sizes, ranging from $256 \times 256$, $512 \times 512$, $768 \times 768$, and $1024 \times 1024$, and on the whole images. 
As the results summarized in Figure~\ref{fig:pvse}, the IndexNet method without context aggregation modules exhibits a monotonically decreasing error trend. In contrast, the matting methods with context aggregation modules experience a reduction in errors initially as the patch size increases, followed by a subsequent increase or stabilization. This observation contradicts our understanding and suggests that both the encoder-decoder network and context aggregation modules help aggregate contexts. 
However, it is evident that context aggregation modules are highly sensitive to the variations in context scale due to the differences in image sizes between the training and inference phases.
This sensitivity proves detrimental to the performance of matting networks employing such modules.

\noindent \textbf{Patch-based Training.}
Matting networks learn to aggregate context information from the data, during the training phase.
The context aggregation modules in the network, with a larger receptive field compared to the network layers in the encoder-decoder, are believed to enhance the utilization of context information for better predictions. 
To validate this understanding, we evaluate matting networks with and without context aggregation modules that are trained on image patches of different sizes. 
Specifically, we evaluated IndexNet without a context aggregation module, and FBAMatting as well as MatteFormer with a context aggregation module. 
All compared methods are first trained on image patches with sizes of $256 \times 256$, $512 \times 512$, $768 \times 768$, and $1024 \times 1024$, and then evaluated on the validation set. 
Note that, we train more epochs for those networks that are trained on smaller image patches.
The results are summarized in Table~\ref{tab:res}. 
Remarkably, we a decrease in error for all networks with an increase in patch size, signifying the advantageous impact of larger training data sizes on matting networks.
Furthermore, the performance of FBAMatting and MatteFormer, both having context aggregation modules, does not show further improvement beyond the training image sizes specified in their papers, which suggests  that the context aggregation modules are limited by manually tuned designs, thereby restricting their universality.

\begin{figure}[!t]
    \includegraphics[width=0.9\linewidth]{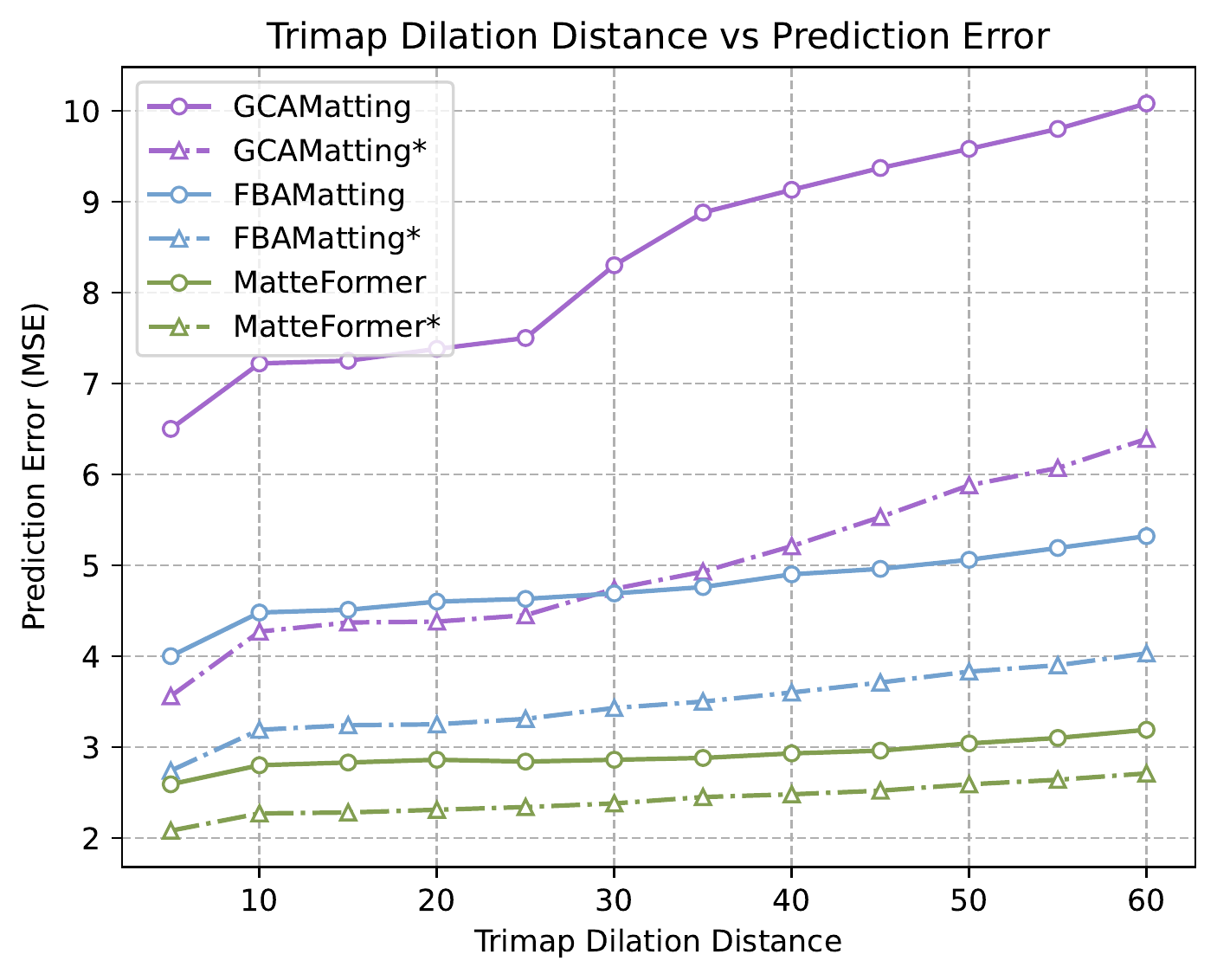}
    \caption{  \textbf{Trimap Dilation Distance} vs \textbf{Prediction Error}. Note that, * denotes the network does not incorporate context aggregation modules.  As the trimap dilation distance increases, the prediction errors (MSE) of all compared matting methods increase. }
    \label{fig:tmse}
\end{figure}

\noindent \textbf{Robustness to Coarse Trimap.}
Recent advancements in matting research~\cite{yu2020mask,dai2022boosting} underscore the importance of robustness to coarse trimaps as a critical performance metric.
To assess the impact of context aggregation modules on handling coarse trimap scenarios, we evaluate existing state-of-the-art matting methods, including GCAMatting, FBAMatting, and MatterFormer, on a modified Adobe Composition-1K dataset featuring trimaps with varying dilation distances. 
The trimap annotations of this dataset are  generated by applying morphological erosion and dilation operations to the ground truth.
Additionally, we evaluate the network variants without context aggregation modules.
The network variants are trained on $1024 \times 1024$ image patches. 
In Figure~\ref{fig:tmse}, we present the results of compared methods, where * denotes network variants without context aggregation modules.
As depicted in the figure, the performance trend of all matting networks consistently degrades as the dilation distance increases, suggesting that the robustness to coarse trimaps is correlated with the encoder-decoder architecture rather than the presence of context aggregation modules.
Furthermore, matting methods with context aggregation modules do not outperform basic networks without such modules, further highlighting  their limited universality due to the sensitivity of context aggregation modules to context scale.

\begin{table}[!t]
  \centering
  \caption{Comparison of the basic matting networks with state-of-the-art matting methods on Adobe Composition-1K. * denotes the backbone adopts the dilated convolution trick.}
  \resizebox{1\linewidth}{!}{
    \begin{tabular}{l|c|c|c|c|c}
    \toprule
    Method & Backbone & SAD   & MSE   & Grad  & Conn \\
    \midrule
    IndexNet~\cite{lu2019indices} & MobileNet & 45.80  & 13.00  & 25.90  & 43.70  \\
    \textbf{BasicNet (Ours)}  & MobileNet & \textbf{30.91} & \textbf{6.73}  & \textbf{13.72}  & \textbf{27.17} \\
    \midrule
    GCAMatting~\cite{li2020natural} & ResNet-34 & 35.28  & 9.00  & 16.90  & 32.50  \\
    A2UNet~\cite{dai2021learning} & ResNet-34 & 32.10  & 7.80 & 16.33 & 29.00 \\
    TIMI-Net~\cite{Liu_2021_ICCV} & ResNet-34 & 29.08  & 6.00  & 11.50  & 25.36  \\
    \textbf{BasicNet (Ours)} & ResNet-34 & \textbf{28.08} & \textbf{5.06}  & \textbf{11.39} & \textbf{24.32} \\
    \midrule
        SIM~\cite{sun2021sim}   & ResNet-50* & 28.00    & 5.80   & 10.8  & 24.80 \\
    FBAMatting~\cite{forte2020fbamatting} & ResNet-50* & 26.40  & 5.40   & 10.6  & 21.50 \\
        \textbf{BasicNet (Ours)} & ResNet-50 &\textbf{23.82} & \textbf{4.27}  & \textbf{8.08}   & \textbf{19.02} \\
    \midrule
        Transmatting~\cite{cai2022TransMatting} & Swin-Tiny &26.83 &5.22& 10.62& 22.14  \\
    MatteFormer~\cite{park2022matteformer} & Swin-Tiny & 23.80  & 4.03  & 8.68  & 18.90 \\
    \textbf{BasicNet (Ours)} & Swin-Tiny & \textbf{19.72}  & \textbf{2.97}  & \textbf{6.27}  & \textbf{14.43} \\
    \bottomrule
    \end{tabular}%
    }
        \vspace{-0.2cm}
  \label{tab:bks}%
\end{table}%

\subsection{Exploring Basic Matting Networks}
\label{sec:33}
Based on the above experiments, we observe that the encoder-decoder component in matting networks is less sensitive to context scale compared to the context aggregation modules,  indicating better universality.
 To explore the feasibility of building basic matting networks using  encoder-decoder,   we delve into evaluating basic encoder-decoder networks with various configurations.

\noindent \textbf{Performance of Basic Matting Networks.}
We first evaluate the performance of the basic encoder-decoder matting network without context aggregation modules.
Specifically, we adopt the MobileNet~\cite{sandler2018mobilenetv2}, ResNet-34~\cite{he2016deep}, ResNet-50~\cite{he2016deep}, and Swin-Tiny~\cite{liu2021Swin} backbones to construct basic matting networks without any context aggregation modules.
Note that, we simply adopt IndexNet as the MobileNet based basic matting network.
Then, we follow the training pipeline of TIMI-Net to train these basic matting networks on image patches with the size of $1024 \times 1024$.
Finally, we compare these basic networks with state-of-the-art networks including, IndexNet, GCAMatting, FBAMatting, A2UNet~\cite{dai2021learning}, TIMI-Net, FBAMAtting, and MatteFormer.
As shown in Table~\ref{tab:bks},  the   basic  matting networks (referred to as BasicNet) outperform state-of-the-art methods, which suggests  the feasibility of building basic matting networks using  encoder-decoder.
Furthermore, the Swin-Tiny and ResNet-50 based networks outperform the MobileNet and ResNet-34   based networks, which suggests that basic matting networks with a larger receptive field may learn better context aggregation to achieve higher performance.

\begin{figure}[!t]
    \begin{center}
    \includegraphics[width=1.\linewidth]{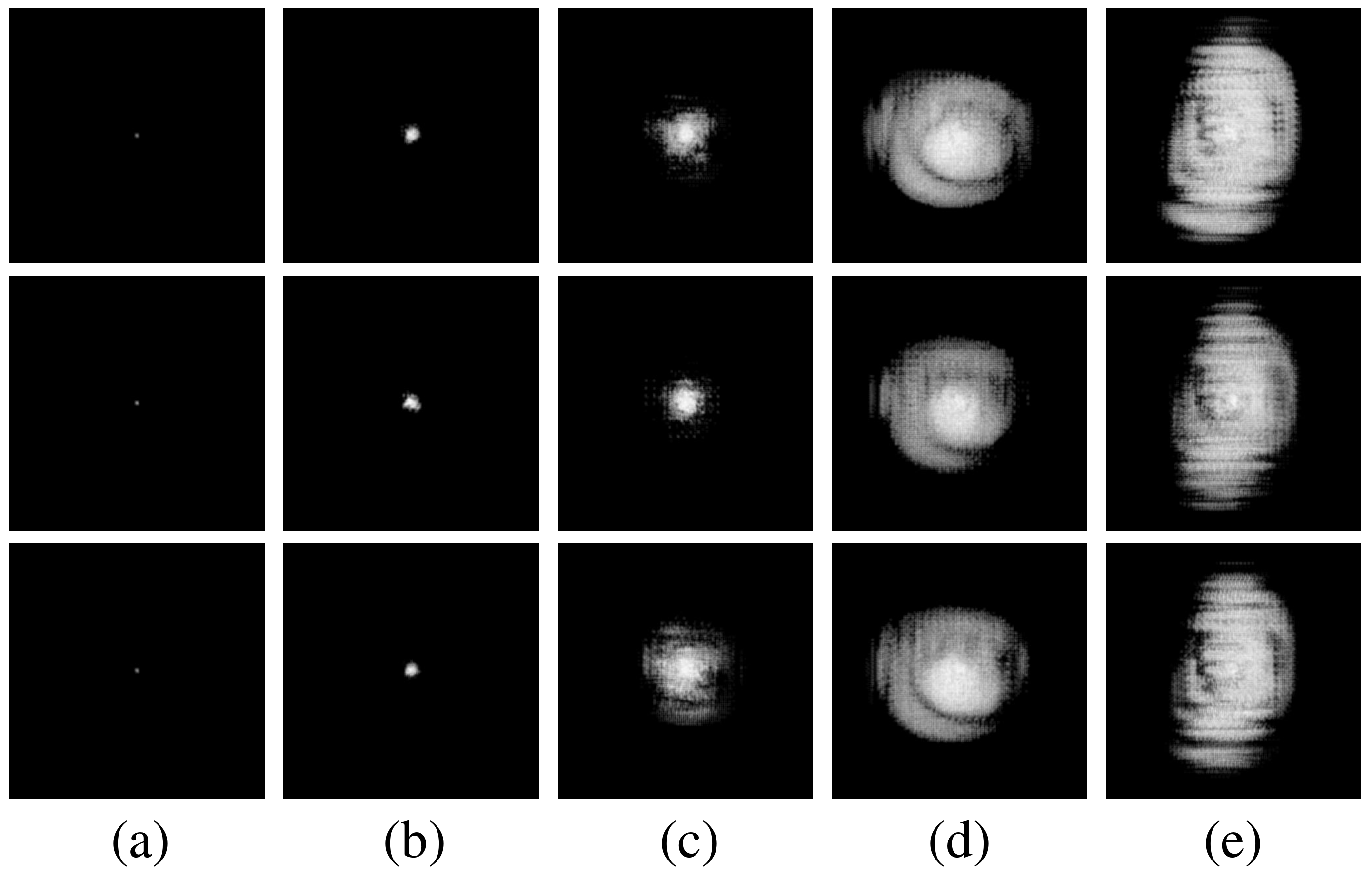}
       \end{center}
            \vspace{-15pt}
    \caption{ Visualization of the receptive field of matting networks trained on image patches of different sizes. (a) Untrained network. (b) Network trained on $256 \times 256$ patches. (c) Network trained on $512 \times 512$ patches. (d) Network trained on $768 \times 768$ patches. (e) Network trained on $1024 \times 1024$ patches. }
    \label{fig:rf}
    \vspace{-5pt}
\end{figure}

\begin{table}[!t]
  \centering
  \caption{Experiment on the training image patch sizes. }
     \resizebox{\linewidth}{!}{
    \begin{tabular}{l|c|cccc}
    \toprule
    Backbone & Patch Size & SAD   & MSE   & Grad  & Conn \\
    \midrule
    Resnet-34~\cite{he2016deep} & $256$    & 41.74 & 12.51  & 22.51 & 40.14 \\
    Resnet-34~\cite{he2016deep} & $512$     & 33.16 & 7.08 & 15.27 & 29.80 \\
    Resnet-34~\cite{he2016deep} & $768$     &  27.70     & 5.41      & 11.23   & 23.89 \\
    Resnet-34~\cite{he2016deep} & $1024$     & 28.08 & 5.06  & 11.39 & 24.32  \\
        \midrule
    Swin-Tiny~\cite{liu2021Swin} & $256$    & 27.99 & 5.30  & 11.23 & 23.96 \\
    Swin-Tiny~\cite{liu2021Swin} & $512$     & 22.42& 3.72 &7.46 & 17.54 \\
    Swin-Tiny~\cite{liu2021Swin} & $768$     & 20.37     & 2.96    & 6.55   & 16.89 \\
    Swin-Tiny~\cite{liu2021Swin} & $1024$     & 19.72 & 2.97  & 6.27 &14.43  \\ 
    \bottomrule
    \end{tabular}%
    }
  \label{tab:drr}%
\end{table}%

\noindent \textbf{Training Image Patch Sizes.}
In our previous experiments on existing matting methods, we observe that matting networks trained with larger image patches may achieve better performance. To explore whether basic matting networks can benefit from large training images,   we train the ResNet-34~\cite{he2016deep} and Swin-Tiny~\cite{liu2021Swin} based basic matting networks with image patches of various sizes, including $256 \times 256$, $512 \times 512$, $768 \times 768$, and $1024 \times 1024$. Subsequently, we evaluate the performance of these networks.
The results,  presented in Table~\ref{tab:drr}, confirm that the performance of matting networks improves with larger training image patches, providing empirical backing for our hypothesis.
To delve deeper into the impact of training image patch sizes on matting networks, we employ the methodology proposed by Luo et al.\cite{luo2016understanding} to visualize the effective receptive field of ResNet-34 based networks trained on image patches of different sizes using gradient feedback, as shown in Figure\ref{fig:rf}. 
The visualization demonstrates that basic matting networks can learn enhanced context aggregation from large image patches.

\noindent \textbf{Receptive Field of Network Layers.}
In our assessment of basic matting networks, we observe a positive correlation between larger receptive fields and improved network performance. 
This observation leads us to hypothesize that the context aggregation capability of a network is positively correlated with its receptive field size. 
To verify this hypothesis, we compare the performance of basic matting networks with different kernel sizes.
Specifically, we build basic matting networks with ResNet-34 and ResNet-50 backbones. 
Then, we replace half of $3 \times 3$  convolutions in these networks with $1 \times 1$ convolutions and  $5 \times 5$  convolutions to control the receptive field.
Finally, we evaluate the modified networks and summarize the results in Table~\ref{tab:arf}. 
The results indicate that matting networks with larger convolution kernels achieve better performance, providing evidence that supports our hypothesis that networks with larger receptive fields exhibit enhanced context aggregation capability.

\subsection{Experimental Findings}
Based on the above  results, we distill   two insights to help design effective matting networks:
(1). Due to manual designs, context aggregation modules are sensitive to changes in context scale, leading to a lack of universality.
(2). Basic encoder-decoder networks possess the capability to learn universal context aggregation. This capability can be further enhanced through training with large image patches and incorporating network layers with a large receptive field.

\begin{table}[!t]
  \centering
  \caption{Experiment on the convolution kernel sizes. } 
   \resizebox{\linewidth}{!}{
    \begin{tabular}{l|c|cccc}
    \toprule
    Backbone & Kernel Size & SAD   & MSE   & Grad  & Conn \\
    \midrule
    Resnet-34~\cite{he2016deep} & $1 \times 1$   & 31.28 & 6.14  & 13.41 & 28.05 \\
    Resnet-34~\cite{he2016deep} & $3 \times 3$     & 28.08 & 5.06  & 11.39 & 24.32 \\
    Resnet-34~\cite{he2016deep} & $5 \times 5$     &  26.72     &  4.74     &  10.08     &22.75 \\
    \hline
    Resnet-50~\cite{he2016deep} & $1 \times 1$    & 28.70  & 5.79  & 10.96 & 24.98 \\
    Resnet-50~\cite{he2016deep} & $3 \times 3$     & 23.82 & 4.27  & 8.08  & 19.02 \\
    Resnet-50~\cite{he2016deep} & $5 \times 5$     & 23.34 & 3.92  & 7.42  & 18.89 \\
    \bottomrule
    \end{tabular}%
}
    \vspace{-0.3cm}
  \label{tab:arf}%
\end{table}%

\begin{figure*}[!t]
    \begin{center}
    \includegraphics[width=0.95\linewidth]{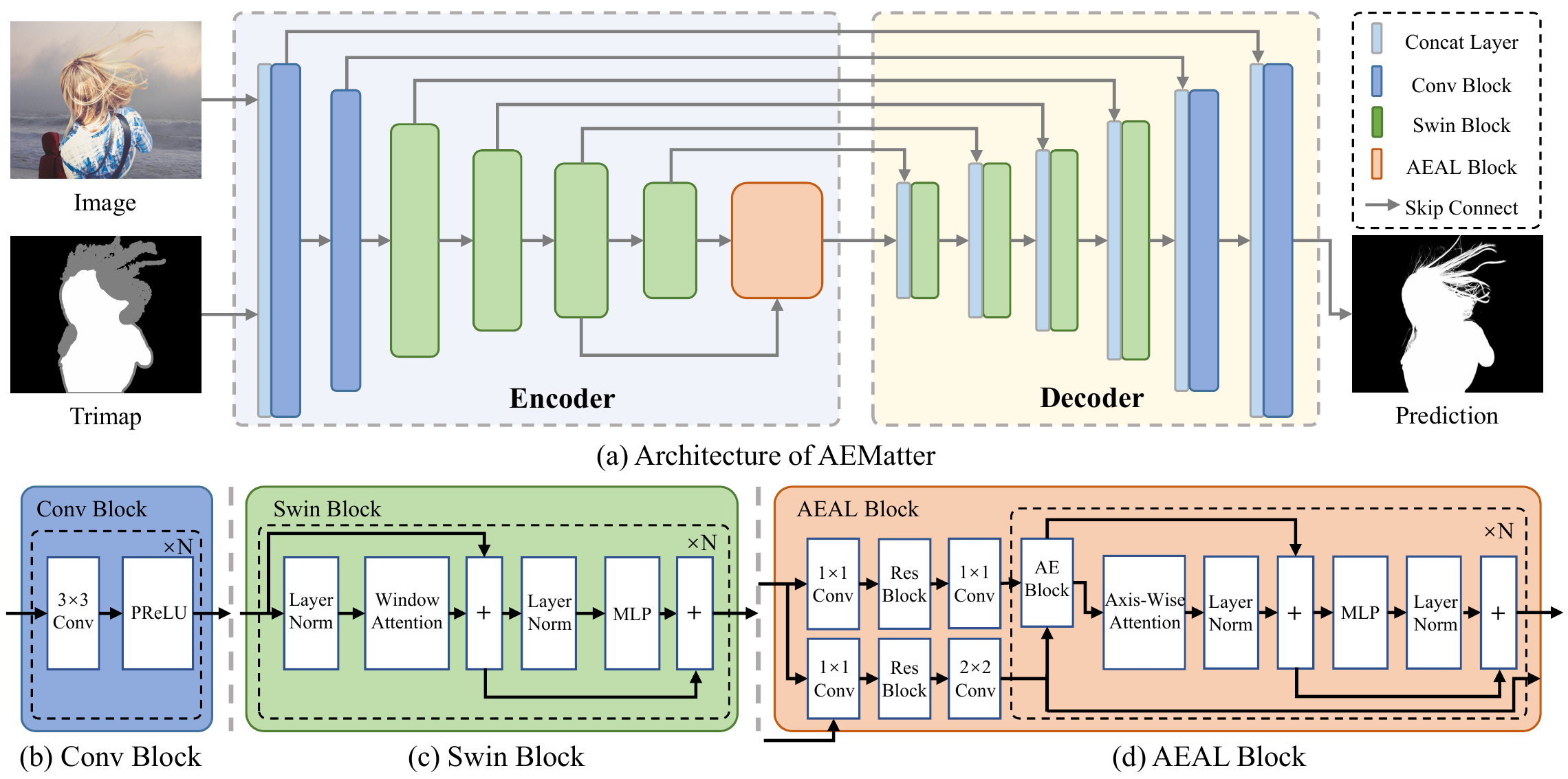}
    \end{center}
    \vspace{-0.4cm}
    \caption{Overview of AEMatter. 
    The encoder adopts a Hybrid-Transformer backbone with appearance-enhanced axis-wise learning blocks to extract context features.
    The decoder adopts Swin blocks to refine the context features and estimate the alpha matte. }
    \label{fig:arch}
\end{figure*}

\section{Proposed Method}
Based on our findings,  we present a simple yet effective matting network, named AEMatter. AEMatter adopts a Hybrid-Transformer backbone with appearance-enhanced axis-wise learning blocks to build a basic  network with strong context aggregation learning capability, as illustrated in Figure~\ref{fig:arch}.
Additionally, AEMatter leverages a large image training strategy to help learn context aggregation.

\subsection{Encoder}
To extract low-level features and context features from the inputs and enlarge the receptive field of AEMatter, we adopt a Hybrid-Transformer backbone with  appearance-enhanced axis-wise learning blocks to construct the encoder.

\noindent \textbf{Hybrid-Transformer Backbone.}
Although the Swin-Tiny~\cite{liu2021Swin} based matting network performs best in the above experiments, Swin-Tiny is primarily  designed for high-level semantic tasks and ignores extracting low-level features, which limits its effectiveness in image matting. 
Prior studies~\cite{park2022matteformer,dai2022boosting} address this issue by incorporating additional shortcut modules to extract low-level features, but their backbones cannot utilize the shortcut features, resulting in subpar performance. 
In contrast, we replace the patch-embedding stem with convolution blocks to extract rich low-level features.
The structure of the convolution block is illustrated in Figure~\ref{fig:arch}(b).
To preserve the image details, we omit the normalization layers in the stem as they affect the information in local regions, which hurts the matting performance. 
In addition, we incorporate PReLU~\cite{he2015delving} as the activation function, which introduces learnable negative  slopes to facilitate network training.
Afterward, we use the Swin blocks of Swin-Tiny to extract high-level context features.

\noindent \textbf{Appearance-Enhanced Axis-Wise Learning.}
The backbone of AEMatter adopts a hierarchical structure that is effective in capturing and integrating context features across large spatial regions.
However, the receptive field of the Swin blocks adopted is still not large enough to cover high-resolution images, which limits the context aggregation capability of the matting network, resulting in sub-optimal performance.
While one possible solution is to employ many downsampling layers and Swin blocks to extract context features across larger regions, such an approach can hinder the training and increase the risk of overfitting.
To address this issue, we incorporate a few appearance-enhanced axis-wise learning  (AEAL) blocks after the backbone, which leverages an appearance-enhanced (AE) block to facilitate training and axis-wise attention to enlarge the receptive fields.

The structure of the AEAL block is illustrated in Figure~\ref{fig:arch}(d).
To mitigate high computational overheads incurred by the high-dimension context features from the backbone, we use residual blocks and $1 \times 1$ convolutions to produce the compact context features $\bm{F}_c$ from the fourth-stage features $\bm{F}_4$ of the backbone.
Additionally, we use $\bm{F}_4$ to guide the extraction of appearance features from third-stage features $\bm{F}_3$ of the backbone with convolution and residual blocks, generating the context-guided appearance features $\bm{F}_a$. 
Subsequently, we employ three cascaded learning modules to process $\bm{F}_c$ and $\bm{F}_a$. 
To facilitate network training, we first introduce an AE block to generate the appearance-enhanced context features $\bm{F}_{ac}$ with $\bm{F}_c$ and $\bm{F}_a$ as 
\begin{equation}
\label{eq:alpha}
\begin{aligned}
\bm{F}_{ac} =\bm{F}_c + {\rm Conv(Res( Conv(Cat(}\bm{F}_c, \bm{F}_a))))
\end{aligned}
\end{equation}
where $\rm Cat(\cdot,\cdot)$, $\rm Conv(\cdot)$, and $\rm Res(\cdot)$ denote the concatenation, $1 \times 1$ convolution, residual block, respectively.
To capture context features over large regions, we propose axis-wise attention, which divides $\bm{F}_{ac}$ into axis-wise rectangular regions and then applies multi-head self-attention.
Specifically, we first zero-pad $\bm{F}_{ac}$ to a size that is an integer multiple of width $W$ and split the padded feature $\bm{F}_{acp}$ into features $\bm{F}_{acpx}$ and $\bm{F}_{acpy}$ along the channel dimension  as 
\begin{equation}
(\bm{F}_{acpx},\bm{F}_{acpy})=\text{Split}_{\text{Channel}}(\text{Pad}(\bm{F}_{ac}))
\end{equation}
where $\text{Split}_{\text{Channel}}(\cdot)$ and $\text{Pad}(\cdot)$ denote the channel wise splitting and zero padding, respectively.
Next, we further split $\bm{F}_{acpx}$ and $\bm{F}_{acpy}$ into two sets of axis-wise features, applying multi-head self-attention to extract context features over large regions. 
These features are then reassembled to form the refined context feature $\bm{F}_{rc}$ as:
\begin{equation}
\begin{aligned}
\bm{F}_{rc}=&\text{Cat}( \text{MHA}(\text{Split}_{\text{Axis-X}}(\bm{F}_{acpx}(X))),\\
            &\text{MHA}(\text{Split}_{\text{Axis-Y}}(\bm{F}_{acpy}(X))) )
\end{aligned}
\end{equation}
where  $\text{MHA}(\cdot)$ denotes the multi-head attention operation. $\text{Split}_{\text{Axis-X}}(\cdot)$ and $\text{Split}_{\text{Axis-Y}}(\cdot)$ denote the x-axis wise splitting and y-axis wise splitting, respectively.
Finally, we utilize the MLP network as the feed-forward network (FFN) for feature transformation, following the vanilla Transformer~\cite{vaswani2017attention}.

\subsection{Decoder}
To enlarge the receptive field of AEMatter and improve the alpha matte estimation, we adopt a Transformer-based decoder that employs Swin blocks which have a large receptive field to refine the context features from the encoder. 
Specifically, we first concatenate the refined context feature $\bm{F}_{rc}$ with the fourth-stage features $\bm{F}_4$ from the encoder, and apply Swin blocks  to generate the initial decoder feature $\bm{F}_d$. 
We then upsample $\bm{F}_d$ and concatenate it with the features of the corresponding scale of the encoder, and apply another Swin block  for feature refinement. 
This process is repeated three times to obtain the refined decoder features $\bm{F}_{rd}$.
To fuse the image details for alpha matte estimation, we upsample $\bm{F}_{rd}$ and concatenate it with the low-level features extracted by the stem of the encoder, and process it using convolution blocks that omit the normalization layers to prevent the mean or variance of the whole feature map from affecting the estimation in local regions.
We perform this process twice and then use a $3 \times 3$ convolution to predict the alpha matte $\bm{\alpha}$. 
Finally, we clip the predicted alpha matte $\bm{\alpha}$ to the range of 0 to 1.

\subsection{Training Strategy}
In our empirical study, we observe that basic encoder-decoder networks can acquire better context aggregation capability when trained on larger image patches, leading to improved matting performance.
Therefore, we propose to train the AEMatter network on $1024 \times 1024$ image patches, which are larger than the existing methods. 
To help AEMatter learn to predict alpha mattes, we define the loss function as
\begin{equation}
\begin{aligned}
\mathcal{L}_{\alpha} = \mathcal{L}_{l1} + \mathcal{L}_{cb} + \mathcal{L}_{lap}
\end{aligned}
\end{equation}
where  $\mathcal{L}_{l1}$,  $\mathcal{L}_{cb}$, and $\mathcal{L}_{lap}$  are the L1  loss,  Charbonnier L1 loss,  and Laplacian loss, which are defined as
\begin{gather}
\mathcal{L}_{l1} =   \left |\bm{\alpha}-\bm{\alpha}^{gt}\right|\\
\mathcal{L}_{cb} = \frac{1}{|\mathcal{T}^U|}\sum_{{i\in\mathcal{T}^U}}{\sqrt{(\alpha_i-\alpha_i^{gt})^2+\epsilon^2}}\\
\mathcal{L}_{lap} =  \sum_{j}{2^{j}\left |{\rm L}_j(\bm{\alpha})-{\rm L}_j(\bm{\alpha}^{gt})\right|}
\end{gather}
where $\bm{\alpha}$ and $\bm{\alpha}^{gt}$ are the predicted alpha matte and ground truth alpha matte of the input image $\bm{I}$, respectively.
Additionally, we adopt training data augmentation techniques, similar to those employed by FBAMatting and MGMatting, to enhance the matting performance.

\begin{figure*}[th]
\includegraphics[width=1\linewidth]{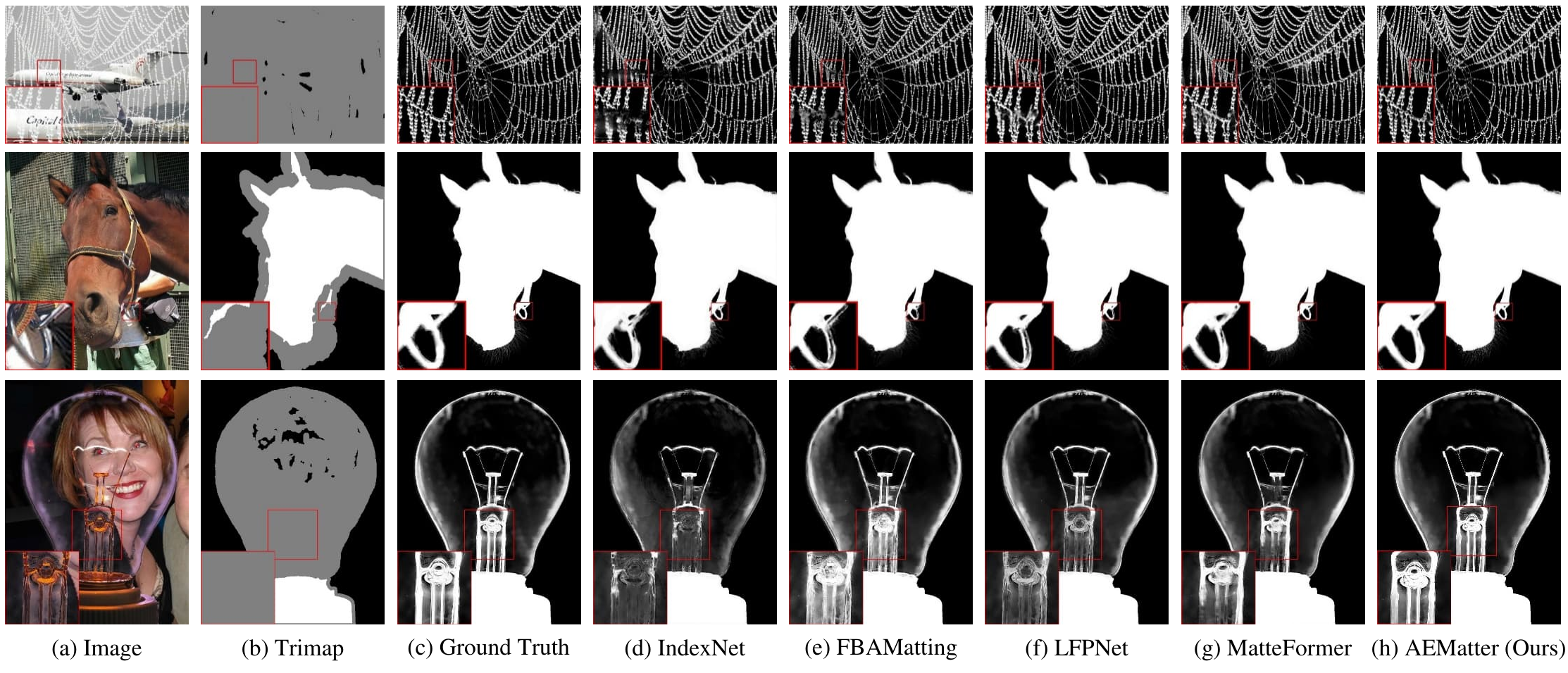}
\vspace{-10pt}
	\caption{Qualitative comparison of the alpha matte results on the Adobe Composition-1K dataset.  }
	\label{fig:adb}
\end{figure*}

\section{Experiments}
In this section, we compare the performance of AEMatter with existing matting methods on the Adobe Composition-1K dataset. 
Additionally, we evaluate the generalization ability of AEMatter on the Distinctions-646~\cite{2020Attention}, Transparent-460~\cite{cai2022TransMatting}, Semantic Image Matting~\cite{sun2021sim}, and Automatic Image Matting-500~\cite{Ijcai2021li} datasets. 
More experimental results and ablation studies are provided in the appendix.

\begin{table}[!t]
  \centering
  \caption{Quantitative results on Adobe Composition-1K. TTA denotes the method adopts the test-time augmentation trick.
  }
  \resizebox{1\linewidth}{!}{
    \begin{tabular}{l|c|c|c|c}
        \toprule
    Method & {SAD} & {MSE} & {Grad} & {Conn} \\
    \midrule
    DIM~\cite{xu2017deep} &50.40  & 17.00  & 36.70  & 55.30  \\
    IndexNet~\cite{lu2019indices}& 45.80  & 13.00  & 25.90  & 43.70   \\
    GCAMatting~\cite{li2020natural} & 35.28  & 9.00  & 16.90  & 32.50  \\
    TIMI-Net~\cite{Liu_2021_ICCV}& 29.08  & 6.00  & 11.50  & 25.36   \\
    SIM~\cite{sun2021sim}  &27.70  & 5.60  & 10.70  & 24.40  \\
    FBAMatting~\cite{forte2020fbamatting} & 26.40  & 5.40  & 10.60  & 21.50  \\
    TransMatting~\cite{cai2022TransMatting}& 24.96 &4.58& 9.72& 20.16  \\
    LFPNet~\cite{liu2021lfpnet} & 23.60  & 4.10  & 8.40  & 18.50    \\
    MatteFormer~\cite{park2022matteformer}& 23.80  & 4.03  & 8.68  &18.90  \\
    dugMatting~\cite{pmlrv202wu23y} & 23.40& 3.90 &7.20& 18.80 \\
        DiffusionMat~\cite{xu2023diffusionmat}& 22.80 &4.00& 6.80 &18.40\\
            DiffMatte-ViTS~\cite{hu2023diffusion}& 20.52&	3.06&	7.05	&14.85  \\
    ViTMatte-B~\cite{yao2024vitmatte}& 20.33& 3.00& 6.74& 14.78 \\
        \midrule
    AEMatter (Ours)  & 17.53&2.26&4.76&12.46\\
    AEMatter + TTA (Ours) &\textbf{16.89}&\textbf{2.06}&\textbf{4.24}&\textbf{11.72} \\ 
        \bottomrule
    \end{tabular}%
  \label{tab:adb}%
 }
\end{table}%

\begin{table}[!t]
  \centering
  \caption{Generalization results on Distinction-646. All methods are trained on Adobe Composition-1K.} 
   \resizebox{1\linewidth}{!}{
    \begin{tabular}{l|c|c|c|c}
        \toprule
    Method & {SAD} & {MSE} & {Grad} & {Conn} \\
    \midrule
    DIM~\cite{xu2017deep} &63.88 & 25.77 & 53.23 & 66.31 \\ 
    IndexNet~\cite{lu2019indices} & 44.93 &9.23 & 41.30 &44.86\\ 
    TIMI-Net~\cite{Liu_2021_ICCV} & 42.61 & 7.75& 45.05 & 42.40 \\ 
    GCAMatting~\cite{li2020natural} & 36.37 &8.19 & 32.34 &36.00\\
    FBAMatting~\cite{forte2020fbamatting} & 32.28 &5.66 & 25.52 & 32.39 \\
        LFPNet~\cite{liu2021lfpnet} & 22.36 & 3.41& 14.92 & 20.50 \\
    Matteformer~\cite{park2022matteformer} & 23.60 & 3.12 & 13.56 & 21.56 \\ 
        \midrule
    AEMatter (Ours)   &  \textbf{16.95}&\textbf{1.81}&\textbf{8.28}&\textbf{14.83} \\   
    \bottomrule
    \end{tabular}%
  \label{tab:d646}%
}
\end{table}%

\subsection{Results on Adobe Composition-1K}
We compare AEMatter against state-of-the-art methods, such as  MatteFormer~\cite{park2022matteformer},  dugMatting~\cite{pmlrv202wu23y},and ViTMatte-B~\cite{yao2024vitmatte} on the Adobe Composition-1K dataset. 
Table~\ref{tab:adb} and Figure~\ref{fig:aim} summarize the quantitative and qualitative results of all compared methods. TTA denotes the method that adopts the test-time augmentation trick.
Quantitative results show that AEMatter significantly outperforms state-of-the-art methods in terms of SAD, MSE, Grad, and Conn metrics. 
Furthermore, qualitative results show AEMatter delivers a visually appealing alpha matte, especially in regions where the foreground and background colors are similar.

\subsection{Generalization on Various Datasets}
To evaluate the generalization ability of AEMatter, we compare AEMatter against existing matting methods on the Distinctions-646, Transparent-460, Semantic Image Matting, and Automatic Image Matting-500 datasets.
It should be noted that all compared matting methods are pre-trained on the Adobe Composition-1K dataset for fair comparison.
We evaluate all compared methods and summarized the quantitative results in Tables~\ref{tab:d646}, \ref{tab:460}, \ref{tab:simd}, and \ref{tab:aim}.
The quantitative results underscore the significant performance advantages of AEMatter compared to existing methods, indicative of its exceptional generalization ability. 

\begin{table}[!t]
  \centering
  {
  \caption{Generalization results on Transparent-460. All methods are trained on Adobe Composition-1K.}
    \resizebox{1\linewidth}{!}{
    \begin{tabular}{l|c|c|c|c}
    \toprule
    Method & {SAD} & {MSE} & {Grad} & {Conn} \\
    \midrule
    DIM~\cite{xu2017deep} &356.20& 49.68& 146.46 &296.31\\
    IndexNet~\cite{lu2019indices} & 434.14 &74.73 &124.98& 368.48\\
    TIMI-Net~\cite{Liu_2021_ICCV} & 328.08& 44.20 &142.11& 289.79\\
    MGMatting~\cite{yu2020mask} &344.65 &57.25 &74.54& 282.79\\
    TransMatting~\cite{cai2022TransMatting} & 192.36& 20.96& 41.80 &158.37\\
    \midrule
    AEMatter (Ours)    & \textbf{122.27}&\textbf{6.92}&\textbf{27.42}&\textbf{112.02} \\ 
    \bottomrule
    \end{tabular}%
  \label{tab:460}%
  }}
\end{table}%

\begin{table}[!t]
  \centering
  \caption{Generalization results on Semantic Image Matting. All methods are trained on Adobe Composition-1K.} 
 \resizebox{1\linewidth}{!}{
    \begin{tabular}{l|c|c|c|c}
        \toprule
    Method & {SAD} & {MSE} & {Grad} & {Conn} \\
    \midrule
    DIM~\cite{xu2017deep} & 95.96&54.25&29.84&100.65 \\ 
    IndexNet~\cite{lu2019indices} & 66.89&25.75&22.07&67.61\\ 
    GCAMatting~\cite{li2020natural} &51.84&19.46&24.16&51.98\\  
    FBAMatting~\cite{forte2020fbamatting} & 26.87 &5.61 & 9.17 & 22.87 \\
    TIMI-Net~\cite{Liu_2021_ICCV} & 54.08& 16.59 &18.91& 53.79\\ 
    LFPNet~\cite{liu2021lfpnet} & 23.05&4.28&23.30&18.19 \\ 
    Matteformer~\cite{park2022matteformer} & 23.90&4.73&7.72&19.01 \\
        \midrule
    AEMatter (Ours)   &\textbf{19.51}&\textbf{2.82}&\textbf{4.62}&\textbf{14.37}\\ %
        \bottomrule
    \end{tabular}%
  \label{tab:simd}%
}
\end{table}%

\begin{table}[!t]
  \centering
  {
  \caption{Generalization results on Automatic Image Matting-500. All methods are trained on Adobe Composition-1K.}
   \resizebox{1\linewidth}{!}{
    \begin{tabular}{l|c|c|c|c}
    \toprule
    Method & {SAD} & {MSE} & {Grad} & {Conn} \\
    \midrule
    DIM~\cite{xu2017deep}& 39.97 & 52.83 & 28.92 & 40.66  \\
    IndexNet~\cite{lu2019indices} & 26.95 & 26.32 & 16.41 & 26.25 \\
    GCAMatting~\cite{li2020natural}& 34.78 & 38.93 & 25.73 & 35.14  \\
    SIM~\cite{sun2021sim}&27.05 &31.10& 23.68& 27.08\\
    FBAMatting~\cite{forte2020fbamatting}& 19.43 & 16.37 & 12.65 & 18.75  \\
    Matteformer~\cite{park2022matteformer}& 26.87 & 29.00    & 23.00    & 26.63   \\
    \midrule
    AEMatter (Ours)   & \textbf{14.76}&\textbf{11.69}&\textbf{11.20}&\textbf{14.19}\\ 
    \bottomrule
    \end{tabular}%
    }
  \label{tab:aim}%
  }
\end{table}%

\section{Conclusion}
In this paper, we revisit the context aggregation mechanisms of matting networks and discover
that discover that a basic encoder-decoder network itself can learn universal context aggregations to achieve high matting performance.
Specifically, we experimentally reveal that while context aggregation modules can effectively aggregate contexts, their sensitivity to context scale restricts the universality.
Simultaneously, we notice that basic encoder-decoder networks can learn context aggregation, leading to impressive matting performance.
Further exploration uncovers that enhancing the context aggregation capability of the network can be achieved through training using large image patches and adopting network layers with a larger receptive field.
Building upon these insights, we introduce a simple yet very effective matting network, named AEMatter, which expands the receptive field of the network with simple structures and is trained using large image patches. 
Experimental results on five datasets demonstrate our AEMatter outperforms state-of-the-art matting methods by a large margin.

\bibliography{ARXIV}
\bibliographystyle{icml2024}


\end{document}